\documentclass{article} 
\usepackage{iclr2022_conference,times}
\iclrfinalcopy

\usepackage{lipsum}
\usepackage[utf8]{inputenc}
\usepackage{amsmath,mathtools,amssymb} 
\usepackage[inline]{enumitem}
\usepackage[detect-weight=true, binary-units=true]{siunitx}
\usepackage{url,hyperref,cleveref}
\usepackage[linesnumbered]{algorithm2e}
\crefname{algocf}{algorithm}{algorithms}
\Crefname{algocf}{Algorithm}{Algorithms}
\usepackage{color,colortbl,etoolbox,xcolor} 
\robustify\bfseries
\usepackage{multirow,booktabs}
\usepackage{subcaption}
\usepackage{csquotes} 
\usepackage[normalem]{ulem} 
\renewcommand\vec{\boldsymbol}

\usepackage{pgfplots}
\usepackage{pgfplotstable}
\pgfplotsset{compat=newest}
\usepgfplotslibrary{groupplots,fillbetween,statistics}
\usetikzlibrary{decorations.pathmorphing,decorations.pathreplacing,arrows.meta,calc,intersections,through,hobby,patterns}

\definecolor{vsrcolor}{RGB}{85, 183, 209}
\usepackage[basecolor=vsrcolor,size=0.4]{vsrs}
\usepackage[thicklines]{cancel}
\newcommand{\U}{U}
\newcommand{\nU}{\cancel{U}}
\newcommand{\D}{D}
\newcommand{\nD}{\cancel{D}}

\pgfplotsset{
    dne/.style 2 args={
        x filter/.append code={
            \edef\tempa{\thisrow{#1}}
            \edef\tempb{#2}
            \ifx\tempa\tempb
            \else
                \def\pgfmathresult{inf}
            \fi
        }
    }
}

\pgfplotstableset{
    /color cells/min/.initial=0,
    /color cells/max/.initial=1000,
    /color cells/textcolor/.initial=,
    color cells/.code={%
        \pgfqkeys{/color cells}{#1}%
        \pgfkeysalso{%
            postproc cell content/.code={%
                \begingroup
                \pgfkeysgetvalue{/pgfplots/table/@preprocessed cell content}\value
                \ifx\value\empty
                    \endgroup
                \else
                \pgfmathfloatparsenumber{\value}%
                \pgfmathfloattofixed{\pgfmathresult}%
                \let\value=\pgfmathresult
                \pgfplotscolormapaccess
                    [\pgfkeysvalueof{/color cells/min}:\pgfkeysvalueof{/color cells/max}]
                    {\value}
                    {\pgfkeysvalueof{/pgfplots/colormap name}}%
                \pgfkeysgetvalue{/pgfplots/table/@cell content}\typesetvalue
                \pgfkeysgetvalue{/color cells/textcolor}\textcolorvalue
                \toks0=\expandafter{\typesetvalue}%
                \xdef\temp{%
                    \noexpand\pgfkeysalso{%
                        @cell content={%
                            \noexpand\cellcolor[rgb]{\pgfmathresult}%
                            \noexpand\definecolor{mapped color}{rgb}{\pgfmathresult}%
                            \ifx\textcolorvalue\empty
                            \else
                                \noexpand\color{\textcolorvalue}%
                            \fi
                            \the\toks0 %
                        }%
                    }%
                }%
                \endgroup
                \temp
                \fi
            }%
        }%
    }
}

\makeatletter
\tikzset{nomorepostaction/.code=\let\tikz@postactions\pgfutil@empty}
\makeatother

\newcommand{\pvalue}[4]{
    \draw ([xshift=.5mm]#1,#3) -- ([yshift=-1mm]#1,#3);
    \draw ([xshift=-.5mm]#2,#3) -- ([yshift=-1mm]#2,#3);
    \draw ([xshift=.5mm]#1,#3) -- node [above,yshift=-.5mm,scale=1] {\tiny{#4}} ([xshift=-.5mm]#2,#3);
}
\newcommand{\pvaluebelow}[4]{
    \draw ([xshift=.5mm]#1,#3) -- ([yshift=1mm]#1,#3);
    \draw ([xshift=-.5mm]#2,#3) -- ([yshift=1mm]#2,#3);
    \draw ([xshift=.5mm]#1,#3) -- node [below,yshift=.5mm,scale=1] {\tiny{#4}} ([xshift=-.5mm]#2,#3);
}
\newcommand{\boxplot}[3]{
    \addplot[black, fill=#3] table[y=#2] {#1};
}
\newcommand{\boxplotalpha}[3]{
    \addplot [black,fill=#3,fill opacity=0.4] table[y=#2] {#1};
}

\newcommand{\boxplotcouple}[4]{
    \boxplot{#1}{#3}{#4}
    \boxplotalpha{#2}{#3}{#4}
}

\definecolor{colorbrewer1}{RGB}{228,26,28}
\definecolor{colorbrewer2}{RGB}{55,126,184}
\definecolor{colorbrewer3}{RGB}{77,175,74}
\definecolor{colorbrewer4}{RGB}{152,78,163}
\definecolor{colorbrewer5}{RGB}{255,127,0}

\title{Collective control of modular soft robots via embodied Spiking Neural Cellular Automata}

\author{%
    Giorgia Nadizar \& Eric Medvet\\
    University of Trieste, Trieste, Italy\\
    \texttt{giorgia.nadizar@phd.units.it,emedvet@units.it}\\
    \And
    Stefano Nichele \& Sidney Pontes-Filho\\
    Oslo Metropolitan University, Oslo, Norway\\
    \texttt{\{stenic,sidneyp\}@oslomet.no}
}

\begin{document}

\maketitle

\begin{abstract}
Voxel-based Soft Robots (VSRs) are a form of modular soft robots, composed of several deformable cubes, i.e., voxels.
Each VSR is thus an ensemble of simple agents, namely the voxels, which must cooperate to give rise to the overall VSR behavior.
Within this paradigm, collective intelligence plays a key role in enabling the emerge of coordination, as each voxel is independently controlled, exploiting only the local sensory information together with some knowledge passed from its direct neighbors (distributed or collective control).
In this work, we propose a novel form of collective control, influenced by Neural Cellular Automata (NCA) and based on the bio-inspired Spiking Neural Networks: the embodied Spiking NCA (SNCA).
We experiment with different variants of SNCA, and find them to be competitive with the state-of-the-art distributed controllers for the task of locomotion.
In addition, our findings show significant improvement with respect to the baseline in terms of adaptability to unforeseen environmental changes, which could be a determining factor for physical practicability of VSRs.
\end{abstract}

\section{Introduction and related works}
\label{sec:introduction}
Biological organisms are intrinsically modular at different scales \citep{lorenz2011emergence}. 
The collective self-organization at the cellular level results in the emergence of complex bodies and brains without any form of centralized control. 
Such modularity allows for mechanisms of local interaction which, in turn, result in collective learning and adaptation. 

In the artificial domain, modular robotics \citep{alattas2019evolutionary} provides a framework for the investigations of biologically-inspired principles of collective control through distributed coordination of the agents composing the robot \citep{cheney2014evolved}. 
In addition, modular robots allow for a high degree of reconfiguration and self-assembly \citep{pathak2019learning}, as well as fault tolerance and modules reusability. 
However, in order to exploit such opportunities, there is the need for modular distributed controllers, possibly embedded in each module. 
Therefore, the overall behavior of the robot is the result of the collective interplay of distributed sensing, local communication, and actuation of interacting body modules. 
In addition, identical modules would facilitate the reusability of the parts and robustness in case of damage \citep{huang2020one}. 
In this work, we focus on a specific type of modular robots, namely Voxel-based Soft Robots (VSRs) \citep{hiller2012automatic}. 
Since VSRs are robots made of interconnected soft blocks (voxels), each module may be considered an agent in a collective. 
As such, mechanisms of collective intelligence are desired. 
While such mechanisms of collective intelligence are rather popular in the context of swarm robotics \citep{hamann2018swarm}, e.g., via self-assembly of thousands of robots through local interactions \citep{rubenstein2014programmable}, they are less explored in the context of modular robotics.

One paradigm of distributed neural control through local interactions of identical cells is Neural Cellular Automata (NCA) \citep{li2002neural,nichele2017neat,mordvintsev2020growing}. 
In case of robots composed of identical modules, each NCA cell can be embodied in a robot module. 
Such approach would in theory facilitate a physical realization as no global wiring nor centralized control would be needed. 
NCA have been successfully used to grow and replicate CA shapes and structures with neuroevolution \citep{nichele2017neat} and with differentiable learning \citep{mordvintsev2020growing}, to produce self-organising textures \citep{niklasson2021self}, to grow 3D artifacts \citep{sudhakaran2021growing}, for regenerating soft robots \citep{horibe2021regenerating}, and for controlling reinforcement learning agents \citep{variengien2021towards}. 


In this work, we introduce a novel embodied NCA model based on more biologically plausible Spiking Neural Network (SNN) controllers. 
We name it embodied Spiking Neural Cellular Automata (SNCA). 
SNNs incorporate neuronal and synaptic states in their neuron models, as well as the concept of time. 
SNCA open up several opportunities in the domain of modular robotics, such as mechanisms of homeostatic adaptation and local learning rules, e.g., spike-timing-dependent plasticity. 
In addition, nearby modules communicate natively through spikes which are not generated at every clock-cycle but only when the internal neural membrane potential reaches a specific threshold, which in turn changes the membrane potential of the post-synaptic neuron, either within the same robotic module or in a nearby module in case of modular robots. 
The advent of neuromorphic hardware, which natively supports SNNs execution and learning, may provide orders of magnitude improvement in energy consumption as compared to traditional neural networks \citep{blouw2019benchmarking}. 
Low energy consumption is considered to be an enabling factor for the physical realization of self-organizing VSRs. 

This work is organized as follows: in \Cref{sec:vsrs} we introduce VSRs, their morphology, and the proposed NCA-based controllers. 
In \Cref{sec:snn} we describe SNNs and SNCA. 
In \Cref{sec:experiments} we present the experimental results and discuss the insights they provide. 
Finally, in \Cref{sec:conclusions} we draw the conclusions.




\section{Collective control of Voxel-Based Soft Robots}
\label{sec:vsrs}
Voxel-Based Soft Robots \citep{hiller2012automatic} are a kind of modular soft robots, composed of several elastically deformable cubes (\emph{voxels}).
In this work, we experiment with a 2D version of VSRs, simulated in discrete time and continuous space \citep{medvet20202d}.
The way in which VSRs achieve movement is a direct consequence of the unique combination of softness and modularity. 
The global behavior derives from the collective and synergistic contraction and expansion of individual voxels, similarly to what happens in biological muscles.
Because of modularity, a VSR can be considered as an ensemble of simple sub-agents, the voxels, which are physically joined to obtain a greater structure, and whose individual behaviors influence eachother and concur to the emergence of coordination.
Therefore, to characterize a VSR we need to specify how the voxels are assembled and what are their properties (\emph{morphology}), and how the voxels compute their control signals and communicate with one another (\emph{controller}).

\subsection{VSR morphology: assembling individual voxels}
\label{sec:vsr-morphology}
The morphology of a VSR specifies how individual voxels are assembled, their sensory equipment, and their physical properties.
A VSR can be represented as a 2D grid of voxels, describing their spatial organization and assembly.
Adjacent voxels in the grid are rigidly linked: not only does this allow to assemble a robot out of primitive modules, but it also forces mutual physical interactions resulting in an overall complex dynamics.
In addition, each voxel can be equipped with sensors to enable proprioception and awareness of the surroundings.
For each voxel, we use three types of sensors:
\begin{enumerate*}[label=(\alph*)]
    \item \emph{area} sensors, perceiving the ratio between the current area of the voxel and its rest area,
    \item \emph{touch} sensors, sensing if the voxel is in contact with the ground or with another body, and
    \item \emph{velocity} sensors, which perceive the velocity of the center of mass of the voxel along the $x$- and $y$-axes.
\end{enumerate*}
We normalize sensor readings to be defined in $[0,1]^4$.


Concerning physical properties, we model voxels as compounds of spring-damper systems, masses, and distance constrains \citep{medvet2020design}, whose parameters can be changed to alter features like elasticity or actuation power.
Each voxel changes volume (actually area, in the 2D case) over time, due to passive interactions with other bodies and the active application of a control signal.
The latter is computed at each simulation time step and is defined in $[-1,1]$, where $-1$ corresponds to maximum requested expansion and $1$ corresponds to maximum requested contraction.
In the employed model \citep{medvet20202d}, contraction and expansion correspond to linear variations of the rest-length of the spring-damper system, proportional to the control signal received.

\subsection{VSR controller: the embodied Neural Cellular Automata paradigm}
\label{sec:vsr-controller}
A VSR controller derives from the ensemble of individual voxels controllers.
However, achieving coordination while keeping the intelligent control of each voxel fully independent of the others is a difficult task.
In fact, most studies involving VSRs either rely on independent, yet not intelligent, controllers based on trivial sinusoidal functions \citep{hiller2012automatic,corucci2018evolving,kriegman2018morphological}, or sacrifice modularity and deploy a central neural controller that has access to all voxels \citep{talamini2019evolutionary,ferigo2021evolving,nadizar2021effects,nadizar2022merging}.
To solve this issue, \citet{medvet2020evolution} introduced the concept of distributed neural controllers, which exploit message passing between neighbors to allow the emerge of coordination thanks to collective intelligence.
Here, we follow along the same direction, combining the key ideas of modularity and collective intelligence, with an approach based on Neural Cellular Automata (NCA) techniques \citep{li2002neural,nichele2017neat,mordvintsev2020growing}, in which the lookup table of each Cellular Automaton (CA) cell is replaced by an Artificial Neural Network (ANN).
More in detail, we consider each voxel as a single \emph{embodied} NCA cell (from now on, simply referred to as NCA), which has access to the local sensor readings and to some information coming from the neighbors, to compute the local actuation value and the messages directed towards adjacent voxels.
Our approach, anyhow, has a substantial difference with the standard NCA architectures: namely, it is strongly bound to the VSR morphology employed, as we only instantiate NCA cells in correspondence to the voxels.

Formally, at every time step $k$, each NCA takes as input a vector $ \vec{x}^{(k)} = \left[ \vec{r}^{(k)} \ \vec{i}_{\uparrow}^{(k)} \ \vec{i}_{\leftarrow}^{(k)} \ \vec{i}_{\downarrow}^{(k)} \ \vec{i}_{\rightarrow}^{(k)} \right]$ and produces as output a vector $\vec{y}^{(k)} = \text{ANN}_{\vec{\theta}}\left(\vec{x}^{(k)}\right) =\left[ a^{(k)} \ \vec{o}_{\uparrow}^{(k)} \ \vec{o}_{\leftarrow}^{(k)} \ \vec{o}_{\downarrow}^{(k)} \ \vec{o}_{\rightarrow}^{(k)}\right]$ where $\vec{r}^{(k)} \in \mathbb{R}^4$ is the local sensor reading, $\vec{i}_{\uparrow}^{(k)}, \vec{i}_{\leftarrow}^{(k)}, \vec{i}_{\downarrow}^{(k)}, \vec{i}_{\rightarrow}^{(k)}$, each defined in $\mathbb{R}^{n_c}$, are values coming from adjacent voxels (from above, left, below, right, respectively, and set to $\vec{0} \in \mathbb{R}^{n_c}$ if no voxel is present in the corresponding direction), $a^{(k)}$ is the actuation value, $\vec{o}_{\uparrow}^{(k)}, \vec{o}_{\leftarrow}^{(k)}, \vec{o}_{\downarrow}^{(k)}, \vec{o}_{\rightarrow}^{(k)}$, each defined in $\mathbb{R}^{n_c}$, are values directed to adjacent voxels (to above, left, below, right, respectively), and $\vec{\theta}$ are the parameters of the ANN constituting the NCA.
Values output by a NCA at time $k$ are used by an adjacent NCA at time $k+1$, e.g., given a NCA $a$ that outputs $\vec{o}_{a,\rightarrow}^{(k)}$, the NCA $b$ at its right will have $\vec{i}_{b,\leftarrow}^{(k+1)}=\vec{o}_{a,\rightarrow}^{(k)}$.

We experiment with three ways of instantiating the general scheme described above, \emph{non-uniform directional} (\nU{}\D{}), \emph{uniform directional} (\U{}\D{}), and \emph{uniform non-directional} (\U{}\nD{}), which differ in the homogeneity of the individual cells (uniform vs.\ non-uniform) and in the information passed between voxels (directional vs.\ non-directional).
The first two schemes are inspired by already existing forms of distributed controllers \citep{medvet2020evolution} and we consider them as baselines, whereas the \U{}\nD{}-NCA is novel.

The most evident, yet conceptually simple, difference lies in the \emph{uniformity}: in \nU{}-NCA, cells have a different ANN in each voxel (each with parameters $\vec{\theta}_i$), whereas in \U{}-NCA all cells ANNs share the same parameters $\vec{\theta}$.
Therefore, it follows that, for a given ANN architecture, the amount of parameters of \nU{}-NCA is $n_{\text{voxels}}$ times the amount of parameters of \U{}-NCA.

The second distinguishing element is \emph{directionality}.
In \nD{}-NCA, cells send the same output to all the adjacent cells, i.e., $\vec{o}_{\uparrow}^{(k)}=\vec{o}_{\leftarrow}^{(k)}=\vec{o}_{\downarrow}^{(k)}=\vec{o}_{\rightarrow}^{(k)}=\vec{o}^{(k)}$ and $\vec{y}^{(k)}=\left[ a^{(k)} \ \vec{o}^{(k)}\right]$, whereas in \D{}-NCA, cells send, in general, different outputs.
The \D{}-NCA hence corresponds to the one originally proposed by \citet{medvet2020evolution}.
Contrarily, \nD{}-NCA are more adherent to the original concept of NCA as $\vec{o}^{(k)}$ can be interpreted as the current state of the cell.

The proposed types of NCA controllers can be employed with any type of ANN, either fully-connected feed-forward NNs, i.e., multi-layer perceptrons (MLP), or more biologically plausible NNs, such as the SNNs described in the following section.

\section{Spiking Neural Networks as robotic controllers}
\label{sec:snn}
Spiking Neural Networks (SNNs) are a type of ANNs in which biological resemblance plays a fundamental role \citep{gerstner2002spiking}.
Often referred to as the third generation of ANNs \citep{maass1997networks}, SNNs are characterized by a more biologically and neuroscientifically faithful neural model than classical ANNs.

The key element of SNNs is the modeling of the evolution over time of the membrane potential of neurons.
Modifications of such potential are caused by incoming neural stimuli, which can either be excitatory (increasing the potential) or inhibitory (decreasing it).
Neural stimuli occur in the form of spikes over time, which can propagate along synapses in order to reach different neurons of the SNN, enabling information passing within the network.
The generation of said spikes is called \emph{firing}, and arises whenever the membrane potential of a neuron exceeds a given threshold.
Despite the binary nature of spikes, the intensity of any stimulus received by a neuron is modulated by the strength of the synapse connecting the firing neuron (pre-synaptic neuron) and the neuron receiving the spike (post-synaptic neuron).
Not unlike classical MLPs, synapses are modeled as weighted connections, where the weights play the main role in determining the behavior of the ANN, and can be subject to task-oriented optimization.

What is indeed essentially different between MLPs and SNNs, is the way information is encoded, which is a direct consequence of the peculiarities of the two models.
In particular, in MLPs there is no notion of time, and information is encoded in the form of real values traveling along the synapses.
Conversely, SNNs are bound to the concept of time to compute the evolution of the neural membranes and for the propagation of spikes in the network.
Within this framework, information is embedded in the time distribution of spikes. 
Hence, additional tools are required to interpret spike trains as real values and vice versa.

Given their high biological resemblance, SNNs are extremely promising robotic controllers.
In fact, faithfully mimicking the functioning of the nervous systems of living organisms could be an enabling factor for bringing the desirable traits of biological organisms to artificial agents, e.g., autonomy or adaptability.
Moreover, the possibility of deploying SNNs on highly energy efficient neuromorphic hardware \citep{li2014activity} is an additional profitable feature, which could be of paramount importance with reference to energy constraints.

\subsection{Discrete time Leaky Integrate and Fire model}
\label{sec:lif}
Several spiking neuron models have been proposed \citep{izhikevich2004model}, which, despite differing in terms of biological plausibility and computational costs, all share the main concepts derived from neuroscience.
Among them, we employ the computationally efficient Leaky Integrate and Fire (LIF) model, simulated in discrete time.
The LIF model represents the neural membrane as a capacitor, whose potential can be increased or decreased by inputs (excitatory or inhibitory), and exponentially decays with time.
At each neural simulation time step $h$, the membrane potential $v^{(h)}_j$ of a LIF neuron $j$ is updated as:
\begin{equation}
    \label{eq:membrane-potential}
    v^{(h)}_j = v^{(h-1)}_j + \sum_{i=1}^{n} w_{i,j} s_i^{(h)} - \Delta t_h \lambda_v v^{(h-1)},
\end{equation}
with $w_{i,j} \in \mathbb{R}$ being the synaptic weight of the $i$-to-$j$ synapse, $n$ being the number of incoming synapses, $s_i^{(h)}\in\{0,1\}$ carrying pre-synaptic neuron spike, and $\Delta t_h=1/f_h$ being the neural simulation time resolution. 
After the update, and if the membrane potential $v^{(h)}_j$ exceeds a threshold $\vartheta^{(h)}_j$, the neuron $j$ outputs a spike, i.e., $s^{(h)}_j=1$, and the membrane potential is reset to its resting value $v_\text{rest}$, otherwise $s^{(h)}_j=0$.

We enhance the LIF model introducing the biological concept of homeostatic plasticity.
Homeostasis is a self regulatory mechanism present at various sites of living organisms, which aims at re-establishing equilibrium in contrast to strong stimuli that could unbalance a system \citep{turrigiano2004homeostatic}.
In our case, homeostasis operates as a firing rate regulator, acting on the threshold $\vartheta_j^{(h)}$ of neurons, to prevent excessive or too scarce activity:
\begin{equation}
    \label{eq:homeostasis}
    \vartheta^{(h)}_j = \min\left(\vartheta^{(h-1)}_j,\sum_{i=1}^{n} w_{i,j}\right) + \psi_j^{(h)},
\end{equation}
with $\psi_j^{(h)}$ being a parameter updated as:
\begin{equation}
    \psi_j^{(h)}=
    \begin{cases}
        \psi_j^{(h-1)} + \psi_\text{inc} & \text{if $s_j^{(h-1)}=1$}\\
        \psi_j^{(h-1)} - \psi_j^{(h-1)}\lambda_\psi\Delta t_h &\text{otherwise.}
    \end{cases}
\end{equation}

\subsection{The LIF model inside NCA}
\label{sec:rate-coding}
We employ the LIF model described above within the ANN for the NCA in our robots.
We simulate both the robot mechanical models and the LIF SNNs in discrete time: however, we update the simulation of the LIF SNNs with a greater frequency.
Namely, we update the mechanical model with a frequency $f_k=\SI{60}{\hertz}$ (the default value of the 2D-VSR-Sim by \citet{medvet20202d}) and the SNNs with a frequency $f_h=16 f_k\approx\SI{1}{\kilo\hertz}$ (as suggested by \citet{izhikevich2004model}).

In practice, at each $k$, we build a spike train $\left(s^{(16k)},\dots,s^{(16k+15)}\right) \in \{0,1\}^{16}$ to be fed to the SNNs from each element of the sensor reading $\vec{r}^{(k)}$ and we compute the actuation value $a^{(k)}$ considering the spikes emitted by the corresponding SNN output neuron up to $h=16k$.
Concerning the information traveling between pairs of SNNs, we simply copy the spike trains with a delay of $16$ time steps in the SNN simulation, i.e., one time step in the robot simulation, consistently with the description given in \Cref{sec:vsr-controller}.
For performing the sensor reading and actuation value conversions, we take inspiration from rate coding \citep{bouvier2019spiking}, where real values are mapped to a frequencies, which are then used to generate spike trains \citep{wang2008behavior}.

For spike trains to be fed to input neurons corresponding to sensor readings, we set:
\begin{equation}
    s^{(h)} = 
    \begin{cases}
    1 &\text{if } \exists n \in \mathbb{N} \text{ s.t. } h=h_\text{last} + n \left\lfloor \frac{f_h}{ f^{(k)}} \right\rfloor  \\
    0 &\text{otherwise},
    \end{cases}
\end{equation}
where $h_\text{last}$ is the time step of the last spike to the neuron (initially set to $0$) and $f^{(k)}=r^{(k)} (f_\text{max}-f_\text{min})+f_\text{min}$, $r^{(k)} \in [0,1]$ being the element of the sensor reading.
That is, we first linearly scale the scalar input to a frequency $f^{(k)} \in \left[f_\text{min},f_\text{max}\right]$ and then we emit spikes at frequency $f^{(k)}$, i.e., one spike each $\left\lfloor \frac{f_h}{ f^{(k)}} \right\rfloor$ time steps of the SNN simulation.
We set $f_\text{min}=\SI{5}{\hertz}$ and $f_\text{max}=\SI{50}{\hertz}$ for biological plausibility: hence, with the maximum scalar input $r^{(k)}=1$, we emit one spike each $\frac{\SI{1}{\kilo\hertz}}{\SI{50}{\hertz}}=20$ time steps, whereas with the minimum input $r^{(k)}=0$, we emit one spike each $\frac{\SI{1}{\kilo\hertz}}{\SI{5}{\hertz}}=200$ time steps.

For the actuation value, we set:
\begin{equation}\label{eq:output-conv}
    a^{(k)} = 2\left( \frac{1}{f_\text{max}} \frac{f_h}{n_w} \sum_{k' = k - n_w + 1} ^ k \sum_{h=16k'}^{16k'+15} s^{(h)} \right) - 1.
\end{equation}
That is, we count the spikes in the last $n_w$ robot simulation time steps, we linearly scale this value to $[0,1]$ considering the maximum possible frequency $f_\text{max}$, and then we linearly scale to $[-1,1]$.
The reason why we consider $n_w$ robot simulation time steps, instead of just the current one, is to have a better resolution of the actuation value and, hence, a smoother control.
After preliminary experimentation, we set $n_w=5$.

\section{Experiments and results}
\label{sec:experiments}
We performed an extensive experimental campaign to investigate how coordination can emerge from different forms of collective control.
We aimed at evaluating if we could improve the baselines of distributed control for VSRs \citep{medvet2020evolution} with our novel contribution, the \U{}\nD{}-SNCA.
Therefore, we addressed the following research question: \emph{``are \U{}\nD{}-SNCA superior with reference to the baselines of distributed control?''}.
To determine the effectiveness of a collective controller, we deployed it onto a VSR, and optimized its parameters using as quality measure the velocity achieved by the robot performing locomotion on a flat terrain.
In addition, we also assessed the controllers adaptability, by measuring the VSR velocity, after the optimization, on a set of unseen terrains, i.e., terrains not used to optimize the controller parameters.
With the extent of obtaining more general results, we experimented with three different morphologies.

\subsection{\U{}\nD{}-SNCA vs.\ baseline embodied NCA}\label{sec:snca-vs-baseline}
In order to provide an answer to the posed research question, we started by optimizing the parameters of three variants of NCA for each of the three considered morphologies, for a total of nine VSRs optimizations.

Concerning the NCA, we took into consideration the \U{}\D{}- and \nU{}\D{}-NCA as baselines, and we compared them against the \U{}\nD{}-SNCA.
We used a MLP with $\tanh$ as activation function for both baseline NCA, while we equipped the SNCA with a fully-connected feed-forward SNN based on the LIF neural model augmented with homeostasis, with the following parameters: $v_\text{rest}=\SI{0}{\milli\volt}$, $\lambda_v=\SI{0.01}{\per\second}$, $\vartheta^{(0)}=\SI{1}{\milli\volt}$, $\psi^{(0)}=\SI{0}{\milli\volt}$, $\psi_\text{inc}=\SI{0.2}{\milli\volt}$, $\lambda_\psi=\SI{0.01}{\per\second}$.
For both ANNs, we used \num{1} hidden layer, setting its size equal to the size of the input layer.
We set $n_c=1$ for both \U{}\D{}- and \nU{}\D{}-NCA, and $n_c=4$ for the \U{}\nD{}-SNCA, in order to make the sizes of the ANNs output layers equal.
Our choice of NCA hyper-parameters was driven by some exploratory experiments and by previous work involving SNNs \citep{pontes2019conceptual} and NCA \citep{nichele2017neat,mordvintsev2020growing}.

Regarding the morphologies, we experimented with \num{3} VSRs, a biped \vsr[1mm]{4}{3}{1111-1111-1001}, a comb \vsr[1mm]{7}{2}{1111111-1010101}, and a worm \vsr[1mm]{5}{1}{11111}.
We chose these morphologies to test the NCA controllers versatility, because they resemble different forms of living organisms, which take advantage of their diverse body shapes to achieve diversified gaits.
We relied on 2D-VSR-Sim \citep{medvet2020design} for the VSRs simulation, leaving all parameters to their default values.
We made the code for the experiments publicly available at \url{https://github.com/giorgia-nadizar/VSRCollectiveControlViaSNCA}.

To optimize the NCA parameters, we resorted to evolutionary algorithms (EAs) as they can easily overcome the struggles posed by the non-differentiability in SNNs.
In addition, EAs are well suited for ill-posed problems with many local optima, which makes them particularly appropriate for optimizing the parameters of robotic controllers.
In this study, we used a simple form of evolutionary strategy (ES).
At first, the population is initialized with $n_\text{pop}$ individuals, i.e., numerical vectors $\vec{\theta}$, all generated by assigning to each element of the vector a randomly sampled value from a uniform distribution over the interval $[-1,1]$.
Subsequently, $n_\text{gen}$ evolutionary iterations are performed, until reaching a total of $n_\text{evals}$ fitness evaluations. 
On every iteration, the fittest quarter of the population is chosen to generate $n_\text{pop}-1$ children, each obtained by adding values sampled from a normal distribution $N(0,\sigma)$ to each element of the element-wise mean $\vec{\mu}$ of all parents.
The generated offspring, together with the fittest individual of the previous generation, end up forming the population of the next generation, which maintains the fixed size $n_\text{pop}$.
We used the following ES parameters: $n_\text{pop}=36$, $n_\text{evals}=\num{30000}$, and $\sigma=0.35$.
We verified that evolution was in general capable of converging to a solution with the chosen values, despite the different sizes of the search spaces corresponding to each NCA configuration.


We optimized VSRs for the task of \emph{locomotion} on a flat terrain, the goal being traveling as fast as possible along the positive $x$-axis.
We assessed the performance of a VSR by measuring its average velocity $v_x$ along the $x$-axis during a simulation of $\SI{30}{\second}$.
We discarded the first $\SI{5}{\second}$ of each simulation to exclude the initial transitory from the velocity measurements.
We used $v_x$ as fitness measure for selecting the best individuals in the ES.
For each of the \num{9} VSRs resulting from the combination of \num{3} NCA and \num{3} morphologies, we performed \num{10} independent evolutionary optimizations, i.e., with different random seeds, for a total of \num{90} runs.

Besides testing the VSR effectiveness upon parameters optimization, i.e., at the end of evolution, we also appraised their adaptability.
We define a VSR controller as \emph{adaptable}, if it is able to achieve good performance in locomotion in spite of environmental changes.
To evaluate this in practice, we took each optimized VSR and re-assessed it on a set of unseen terrains, i.e., terrains which none of its ancestors ever experienced locomotion on.
In particular, we experimented with the following terrains:
\begin{enumerate*}[label=(\alph*)]
    \item hilly with $6$ combinations of heights and distances between hills,
    \item steppy with $6$ combinations of steps heights and widths,
    \item downhill with $2$ different inclinations, and
    \item uphill with $2$ different inclinations.
\end{enumerate*}
As a result, we re-assessed each individual on a total of $16$ different terrains; we define its adaptability as the average of the $v_x$ on those terrains (each computed in a \SI{30}{\second} simulation, discarding the initial \SI{5}{\second}).

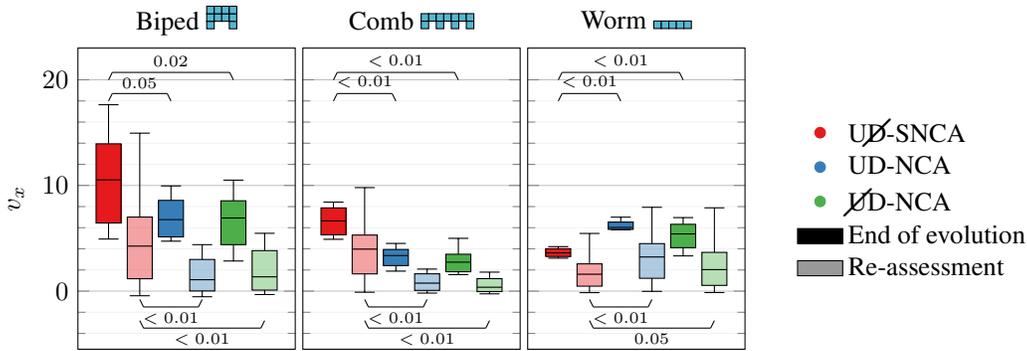
\begin{figure}
    \centering
    \begin{tikzpicture}
        \def\ptopabove{20.7}
        \def\ptopbelow{18.7}
        \def\pbottomabove{-1.5}
        \def\pbottombelow{-3.5}
        \begin{groupplot}[
            boxplot,
            boxplot/draw direction=y,
            width=0.32\linewidth,
            height=0.4\linewidth,
            group style={
                group size=3 by 1,
                horizontal sep=1mm,
                vertical sep=1mm,
                xticklabels at=edge bottom,
                yticklabels at=edge left,
            },
            legend cell align={left},
            ymin=-5.5,ymax=23.0,
            xticklabels=\empty,
            xmajorticks=true,
            xminorticks=false,
            xtick style={draw=none},
            ymajorgrids=true,
            yminorgrids=true,
            minor y tick num=4,
            grid style={line width=.1pt, draw=gray!10},
            major grid style={line width=.15pt, draw=gray!50},
            title style={anchor=center, yshift=1ex},
            legend style={draw=none}
        ]
            \nextgroupplot[
                align=center,
                title={Biped \vsr[1mm]{4}{3}{1111-1111-1001}},
                ylabel={$v_x$}
            ]
            \boxplotcouple{data/boxplot/b_biped_CA4.txt}{data/boxplot/v_biped_CA4.txt}{LIF-H}{colorbrewer1};
            \boxplotcouple{data/boxplot/b_biped_Homo.txt}{data/boxplot/v_biped_Homo.txt}{MLP}{colorbrewer2};
            \boxplotcouple{data/boxplot/b_biped_Hetero.txt}{data/boxplot/v_biped_Hetero.txt}{MLP}{colorbrewer3};
            \pvalue{1}{3}{\ptopbelow}{$0.05$}
            \pvalue{1}{5}{\ptopabove}{$0.02$}
            \pvaluebelow{2}{4}{\pbottomabove}{$<0.01$}
            \pvaluebelow{2}{6}{\pbottombelow}{$<0.01$}
            
            \nextgroupplot[
                align=center,
                title={Comb \vsr[1mm]{7}{2}{1111111-1010101}}
            ]
            \boxplotcouple{data/boxplot/b_comb_CA4.txt}{data/boxplot/v_comb_CA4.txt}{LIF-H}{colorbrewer1};
            \boxplotcouple{data/boxplot/b_comb_Homo.txt}{data/boxplot/v_comb_Homo.txt}{MLP}{colorbrewer2};
            \boxplotcouple{data/boxplot/b_comb_Hetero.txt}{data/boxplot/v_comb_Hetero.txt}{MLP}{colorbrewer3};
            \pvalue{1}{3}{\ptopbelow}{$<0.01$}
            \pvalue{1}{5}{\ptopabove}{$<0.01$}
            \pvaluebelow{2}{4}{\pbottomabove}{$<0.01$}
            \pvaluebelow{2}{6}{\pbottombelow}{$<0.01$}
            
            \nextgroupplot[align=center,
                            title={Worm \vsr[1mm]{5}{1}{11111}},
                            legend style={at={(1.2,.8)},anchor=north west},
                            legend columns=1,
                            legend entries={{\U{}\nD{}-SNCA}, {\U{}\D{}-NCA}, {\nU{}\D{}-NCA}, {End of evolution}, {Re-assessment}},]
            \addlegendimage{mark=*,color=colorbrewer1,fill}
            \addlegendimage{mark=*,color=colorbrewer2,fill}
            \addlegendimage{mark=*,color=colorbrewer3,fill}
            \addlegendimage{area legend,color=black,fill}
            \addlegendimage{area legend,color=black,fill,fill opacity=0.4}
            \boxplotcouple{data/boxplot/b_worm_CA4.txt}{data/boxplot/v_worm_CA4.txt}{LIF-H}{colorbrewer1};
            \boxplotcouple{data/boxplot/b_worm_Homo.txt}{data/boxplot/v_worm_Homo.txt}{MLP}{colorbrewer2};
            \boxplotcouple{data/boxplot/b_worm_Hetero.txt}{data/boxplot/v_worm_Hetero.txt}{MLP}{colorbrewer3};
            \pvalue{1}{3}{\ptopbelow}{$<0.01$}
            \pvalue{1}{5}{\ptopabove}{$<0.01$}
            \pvaluebelow{2}{4}{\pbottomabove}{$<0.01$}
            \pvaluebelow{2}{6}{\pbottombelow}{$0.05$}
            
        \end{groupplot}
    \end{tikzpicture}
    \caption{
        Box plots of the velocities $v_x$ achieved by the best individuals at the end of evolution and upon re-assessment on unseen terrains for different VSR morphologies (plot columns) and embodied NCA controllers (color).
        Above pairs of boxes we report the $p$-values resulting from Mann-Whitney U tests with the null hypothesis of equality of the means.
    }
    \label{fig:snca-nca}
\end{figure}

The results of our experimental evaluation are summarized in \Cref{fig:snca-nca}.
More in detail, for each of the considered VSR morphologies and NCA variants, we display the distribution of velocities achieved at the end of evolution by the best individuals, and their performance in terms of adaptability, i.e., the distribution of their average velocity on unseen terrains.
In addition, above pairs of box plots, we report the $p$-values resulting from a two-sided Mann Whitney U statistical test; we consider, unless otherwise specified, $\alpha=0.05$ as confidence level.

From \Cref{fig:snca-nca}, we observe that for the biped and the comb morphologies, \U{}\nD{}-SNCA are always significantly better than the baseline in terms of adaptability, and they are significantly better at the end of evolution in all but one case.
However, the outcomes seem to be exactly opposite for the worm morphology, making it difficult to provide a general answer to the posed research question.

To further investigate on this apparent contradiction, we examined the behavior of a few evolved VSRs (videos are available at \url{https://giorgia-nadizar.github.io/VSRCollectiveControlViaSNCA/}) and found the reason behind the failure of the SNCA to be glaring for the worm morphology.
In particular, we noticed that the NCA based on MLPs trigger a high frequency dynamics, resulting in a vibrating behavior, which for SNCA is prevented by homeostasis and $n_w=5$ when converting spikes to actuation values.
However, for the worm morphology, vibration appears to be the only effective gait, as these VSRs are not able to properly bend, having only one row of voxels at disposal.
Conversely, the biped and comb morphologies have more complex structures, which allow the discovery of a wider range of efficacious gaits.
In fact, when we inspected the behaviors of these two families of VSRs, we could notice a broader variety of gaits, with some tendencies to vibration among those controlled by MLP-based NCA.
Avoiding vibrating behaviors, which have been shown to be a strong attractor in evolution \citep{medvet2021biodiversity}, is of paramount importance, as this type of movement severely hinders adaptability, and constitutes an insurmountable barrier for the physical practicability of VSRs, i.e., a form of reality gap \citep{van2021influence,salvato2021crossing}.
Even though it is possible to explicitly discourage vibrating behaviors, e.g., decreasing the actuation frequency, having a controller which avoids them by design is an undeniably significant accomplishment.


\subsection{Strengths of the uniform non-directional SNCA}\label{sec:success-snca}
From the experimental outcomes, however, another question arises: \emph{``what are the reasons behind the success of the \U{}\nD{}-SNCA?''}.
Namely, does the improvement lay in the non-directionality of the NCA or in the SNN employed?

To address the newly emerged points, we deepened our analysis with a supplementary experimental campaign, encompassing new combinations of NCA architectures, neural models, and morphologies, for a total of \num{12} additional VSRs to be optimized.
Regarding the morphologies, we experimented with the biped and the comb, discarding the worm for the reasons highlighted in \Cref{sec:snca-vs-baseline}.
For what concerns the controllers, we extended the previous experiments by evaluating all missing combinations of NCA architecture and neural models.
Among the latters, we also included a SNN composed of LIF neurons for which we disabled homeostasis, keeping the value of the threshold fixed throughout the simulation to its initial value $\vartheta_i^{(h)}=\vartheta_i^{(0)}=\SI{1}{\milli\volt}$.
For each of the \num{12} new VSRs we repeated the experimental pipeline of \Cref{sec:snca-vs-baseline}.

We display the results, together with the outcomes of the previous experiments, in \Cref{tab:summary}.
Each cell of the table reports the median of the velocities achieved by VSRs at the end of evolution and upon re-assessment, grouped by morphology; each row corresponds to a NCA architecture, whereas we put neural models on the columns.
We color cells proportionally to the median of velocities in order to better convey the information.

\begin{table}[ht]
    \centering
    \pgfplotstableread[col sep=comma]{data/heatmap/heatmap.txt}\mytable
    \pgfplotstabletypeset[
        color cells={min=0,max=15},
        col sep=comma,
        columns/\space/.style={reset styles,string type,column type = {l},column type/.add={}{@{\hspace{.4em}}}},
        every head row/.style={
            before row={
            \toprule
            & \multicolumn{6}{c}{End of evolution} & \multicolumn{6}{c}{Re-assessment} \\
            \cmidrule(l{-0.5em}r{1em}){2-7} \cmidrule(l{-0.5em}r){8-13}
            & \multicolumn{3}{c}{Biped \vsr[1mm]{4}{3}{1111-1111-1001}} & \multicolumn{3}{c}{Comb \vsr[1mm]{7}{2}{1111111-1010101}} & \multicolumn{3}{c}{Biped \vsr[1mm]{4}{3}{1111-1111-1001}} & \multicolumn{3}{c}{Comb \vsr[1mm]{7}{2}{1111111-1010101}} \\
            \cmidrule(l{-0.5em}r{1em}){2-4} \cmidrule(l{-0.5em}r{1em}){5-7} \cmidrule(l{-0.5em}r{1em}){8-10} \cmidrule(l{-0.5em}r){11-13}
            },
            after row=\midrule,
        },
        every last row/.style={after row=\bottomrule},
        columns/1MLP/.style={column name={M},column type/.add={}{@{\hspace{.7em}}}},
        columns/2MLP/.style={column name={M},column type/.add={}{@{\hspace{.7em}}}},
        columns/3MLP/.style={column name={M},column type/.add={}{@{\hspace{.7em}}}},
        columns/4MLP/.style={column name={M},column type/.add={}{@{\hspace{.7em}}}},
        columns/1LIF/.style={column name={S},},
        columns/2LIF/.style={column name={S},},
        columns/3LIF/.style={column name={S},},
        columns/4LIF/.style={column name={S},},
        columns/1LIF-H/.style={column name={S-H},column type/.add={@{\hspace{.7em}}}{}},
        columns/2LIF-H/.style={column name={S-H},column type/.add={@{\hspace{.7em}}}{}},
        columns/3LIF-H/.style={column name={S-H},column type/.add={@{\hspace{.7em}}}{}},
        columns/4LIF-H/.style={column name={S-H},column type/.add={@{\hspace{.7em}}}{}},
        /pgfplots/colormap={greenish}{
            rgb255=(240,249,232)
            rgb255=(168,221,181)
            rgb255=(123,204,196)
            rgb255=(78,179,211)
            rgb255=(43,140,190)
            rgb255=(8,88,158)
        },
        /pgf/number format/fixed zerofill,precision=1
    ]{\mytable}
    \caption{
        Medians of velocities $v_x$ achieved by the best individuals at the end of evolution and upon re-assessment on unseen terrains for different morphologies.
        We put different NCA architectures on each row, and ANN models on the columns (M stands for MLP, S for SNN without homeostasis, S-H for SNN with homeostasis).
        Cells are colored proportionally to $v_x$.
    }
    \label{tab:summary}
\end{table}

From examining \Cref{tab:summary}, we can investigate on the importance of the two aforementioned factors.
First, to weigh the impact of the NCA architecture, we compare the medians of different rows for each column of the table.
We observe that \U{}\nD{}-NCA are not worse than both \D{}-NCA in all but one case, and they always equal or outperform \D{}-NCA if combined with SNNs.
We speculate this descends from the fact that, especially in absence of agent specialization, i.e., in the case of \U{}-NCA, it is easier for the prototype individual to learn to pass a single message to all its clone neighbors and correctly interpret the information received.
In addition, we highlight that \U{}\nD{}-NCA are less prone to triggering vibrating dynamics by design, and are thus more successful in combination with SNNs, which display and take advantage of the same trait. 

Concerning the importance of the neural model, we note that SNNs, either with or without homeostasis, surpass MLPs in \num{10} out of \num{12} cases.
To better appraise the influence of homeostasis in SNNs, we need to narrow our focus to the re-assessment results, where this neural model leads to neatly superior outcomes in all but one case, confirming its fundamental role in self-regulation and adaptation.
Moreover, we can re-state that SNNs seem to be more naturally suited for being combined with \U{}\nD{}-NCA, as both tend to move away from high frequency non-adaptable behaviors.
Therefore, we can conclude that the superiority of our contribution lies in the successful combination of the novel \U{}\nD{}-NCA architecture with SNNs with homeostasis.

\section{Concluding remarks}
\label{sec:conclusions}
We explored the paradigm of collective control of Voxel-based Soft Robots (VSRs), a form of simulated modular soft robots, appraising the emergence of coordination from the synergistic actuation of individual agents, i.e., the voxels constituting the VSR.
Taking inspiration from NCA, a form of distributed neural control, and from state-of-the-art forms of embodied control of VSRs, we introduced the novel concept of embodied Spiking Neural Cellular Automata (SNCA), in which we used Spiking Neural Networks (SNNs) as elementary units.
To evaluate the performance of the proposed SNCA as a robotic controller, we compared it against the state-of-the-art embodied controllers, optimizing the controller parameters of three different VSRs for the task of locomotion.
Our experimental results show that the SNCA is not only competitive with the pre-existing controllers, but it also leads to significantly more adaptable agents, outperforming their rivals when faced with unforeseen circumstances.
Moreover, we highlight a trend towards less reality-gap prone behaviors in VSRs controlled by SNCA, which paves the way for the physical practicability of such robots.

We believe our contribution can be considered as a starting point for several additional analyses, spanning across diverse research directions.
Concerning SNNs, we plan to experiment with neuroplasticity in the form of unsupervised learning, to the extent of achieving greater generality and increased robustness of controllers \citep{qiu2020towards}.
In addition, we will address the problem of collective control with a cooperative coevolution strategy aimed at optimizing an ensemble of heterogeneous SNCA controllers \citep{potter2000cooperative}.

\bibliography{main.bib}
\bibliographystyle{iclr2022_conference}

\end{document}